\documentclass[review,times]{elsarticle}

\usepackage{amsmath,amssymb}
\usepackage{graphicx}
\usepackage{booktabs}
\usepackage{array}
\usepackage{multirow}
\usepackage{longtable}
\usepackage{tabularx}
\usepackage{caption}
\usepackage{subcaption}
\usepackage{placeins}
\usepackage{hyperref}
\usepackage{url}
\usepackage{xcolor}
\usepackage{enumitem}

\graphicspath{{./}{figures/}}

\hypersetup{
  colorlinks = true,
  linkcolor  = blue!60!black,
  citecolor  = blue!60!black,
  urlcolor   = blue!60!black
}

\newcommand{\tablenote}[1]{\par\smallskip\begin{minipage}{\linewidth}\footnotesize\textit{Note.} #1\end{minipage}}

\journal{Knowledge-Based Systems}

\begin{document}

\begin{frontmatter}

\title{Automated Generation of Complexity-Validated Decision Scenarios Using Large Language Models}

\author[ise]{Abdalla Doleh\corref{cor1}}
\ead{ai5145@wayne.edu}
\author[bus]{Toni Somers}
\ead{aa3808@wayne.edu}
\author[ise]{Ratna Babu Chinnam}
\ead{ai2396@wayne.edu}
\cortext[cor1]{Corresponding author.}

\affiliation[ise]{organization={Department of Industrial \& Systems Engineering, Wayne State University},
  city={Detroit}, state={MI}, postcode={48202}, country={USA}}
\affiliation[bus]{organization={Mike Ilitch School of Business, Wayne State University},
  city={Detroit}, state={MI}, postcode={48202}, country={USA}}

% -------------------------------------------------------
% ABSTRACT
% -------------------------------------------------------
\begin{abstract}
\noindent\textbf{Problem:} Cognitive decision-making research requires diverse, complexity-controlled
scenarios, but manual generation is slow, inconsistent, and biased.

\noindent\textbf{Solution:} We present an automated pipeline that generates structured decision
scenarios with large language models (LLMs) and validates their complexity using a composite
framework derived from the task-complexity frameworks of Wood, Campbell, Liu and Li, and Sweller
\citep{Wood1986,Campbell1988,LiuLi2012,Sweller1994}.

\noindent\textbf{Results:} We evaluated 4,238 scenarios across domains and complexity tiers.
Measurement validation met psychometric standards: inter-LLM agreement across five independent
model families was near-perfect (ICC~$= 0.997$; $\kappa = 0.971$), and known-groups validity
showed large tier separation ($\eta^{2} = 0.587$, all pairwise $p < .001$). Factor analysis
revealed a dominant complexity construct ($\lambda = 0.87$--$0.96$ for three frameworks) with
interactivity forming a weaker secondary dimension ($\lambda = 0.34$). Discriminant validity is
limited by a strong relationship between complexity and text length that persists after
controlling for tier (partial $r = 0.86$); this constrains construct purity while leaving the
instrument's tier-grading function intact. Model-performance
analyses revealed substantial throughput differences across models and a negative association
between throughput and schema pass rate ($r = -0.967$, $p = 0.007$, $n = 5$) that is suggestive of
a speed-quality trade-off but is driven largely by a single high-throughput model and requires a
larger panel to confirm. Llama~4 Maverick generated scenarios fastest (134 vs.\ 25 per minute for
DeepSeek Chat V3) but under-produced complex-tier scenarios, whereas DeepSeek Chat V3 combined
balanced domain coverage with high schema compliance.

\noindent\textbf{Impact:} The complexity measurement system demonstrated strong psychometric
properties (ICC~$= 0.997$ inter-rater consensus, $\eta^{2} = 0.587$ known-groups),
enabling reliable classification of scenarios into Simple, Moderate, and Complex tiers.
These validated scenarios provide the measurement infrastructure required for downstream cognitive
assessment of AI systems.
\end{abstract}


\end{frontmatter}

% -------------------------------------------------------
% 1. INTRODUCTION
% -------------------------------------------------------
\section{Introduction}
\label{sec:intro}

Decision-making research depends on standardized, complexity-controlled scenarios. Traditional
scenario design is labor-intensive, prone to researcher bias, and difficult to scale. Recent
advances in large language models (LLMs) make it possible to generate structured scenarios at
scale, but the scientific utility of these scenarios depends on the validity and reliability of
the complexity measurement used to classify them. Without validated measurement, any downstream
model comparisons or experimental conclusions are uninterpretable \citep{Campbell1988,Wood1986}.

This paper introduces and validates an automated scenario generation and complexity scoring
pipeline for cognitive decision research. We treat complexity as a latent construct expressed
through four established frameworks: Wood's task complexity, Campbell's multiple-path complexity,
Liu and Li's cognitive complexity, and Sweller's element interactivity
\citep{Campbell1988,LiuLi2012,Sweller1994,Wood1986}. These frameworks remain foundational
in contemporary research, with recent work extending their application to human-AI
decision-making \citep{Abdul2023,Stadler2025,Wen2025}, organizational performance
\citep{Stajkovic2025}, and cognitive load measurement \citep{Chen2023,Sweller2019}.
We test whether these frameworks converge on a coherent construct, whether complexity scores
behave as theorized, and whether independent LLM judges agree on scoring outcomes. The central
goal is to validate the measurement system as a reusable instrument for complexity-controlled
scenario generation in cognitive research and AI benchmarking.

This work is distinct from three adjacent lines of research. LLM-as-judge evaluation frameworks
use language models to score the quality of other models' \emph{outputs}, for example in
open-ended dialogue benchmarks \citep{Zheng2023}; they assess response quality rather than
generating complexity-controlled \emph{stimuli} with validated difficulty. Automated benchmark
generation produces large, diverse task sets but optimizes for coverage and discriminative power
rather than psychometric construct validity. Synthetic-data generation likewise emphasizes scale
and diversity without enforcing a theory-grounded complexity structure. The present contribution
treats the complexity of generated scenarios as a \emph{measurement instrument} and subjects that
instrument to formal psychometric validation, convergent, discriminant, known-groups, and
nomological validity together with inter-rater agreement, so that the resulting difficulty tiers
can serve as controlled experimental variables. To our knowledge, this combination of
theory-grounded, multi-framework complexity scoring with end-to-end psychometric validation has
not previously been reported for LLM-generated decision scenarios.

\subsection{Contributions}
\label{sec:contributions}

This paper makes four contributions:
\begin{enumerate}[noitemsep]
  \item An end-to-end pipeline for automated scenario generation, schema validation, and
        multi-framework complexity scoring.
  \item A comprehensive construct validation of automated complexity measurement (convergent,
        discriminant, known-groups, nomological, and inter-rater).
  \item A model-performance evaluation across domain balance, complexity coverage, throughput,
        and failure patterns.
  \item Practical guidance for applying validated scenario generation in cognitive research
        and benchmarking.
\end{enumerate}

\subsection{Research Questions and Hypotheses}
\label{sec:hypotheses}

The study addresses measurement validity before comparing model performance. Confirmatory
hypotheses target convergent validity (H1a), discriminant validity (H1b), known-groups validity
(H1c), inter-LLM consensus (H3), and nomological validity (H1d, H1e). Exploratory analyses examine
model performance and failure patterns (H4--H6; RQ5--RQ6). A small number of originally planned
analyses could not be completed with the available data; their scope and the reasons are noted in
Sections~\ref{sec:design} and~\ref{sec:limitations}.

\begin{description}[noitemsep]
  \item[H1a] (Convergent Validity): Framework scores converge on a latent complexity factor.
  \item[H1b] (Discriminant Validity): Complexity scores are distinct from confounds (word count,
        model identity, domain).
  \item[H1c] (Known-Groups Validity): Complexity scores differ significantly and in the predicted
        direction across Simple, Moderate, and Complex tiers.
  \item[H1d] (Nomological, Consensus): Complexity relates negatively to inter-judge consensus agreement.
  \item[H1e] (Nomological, Feature Count): Complexity relates positively to feature count.
  \item[H3]  (Inter-LLM Consensus): Independent LLM judges produce equivalent complexity scores.
  \item[H4--H6] (Model Performance): Models differ in domain balance, tier coverage, and throughput.
\end{description}

\subsection{Conceptual Framework}
\label{sec:conceptual}

The study uses a two-stage framework: (a)~measurement validation of the automated complexity
scoring instrument, and (b)~model-performance analysis contingent on measurement validity.
If measurement validation fails, model comparisons are not interpretable.

\begin{figure}[htbp]
  \centering
  \includegraphics[width=0.60\textwidth,keepaspectratio]{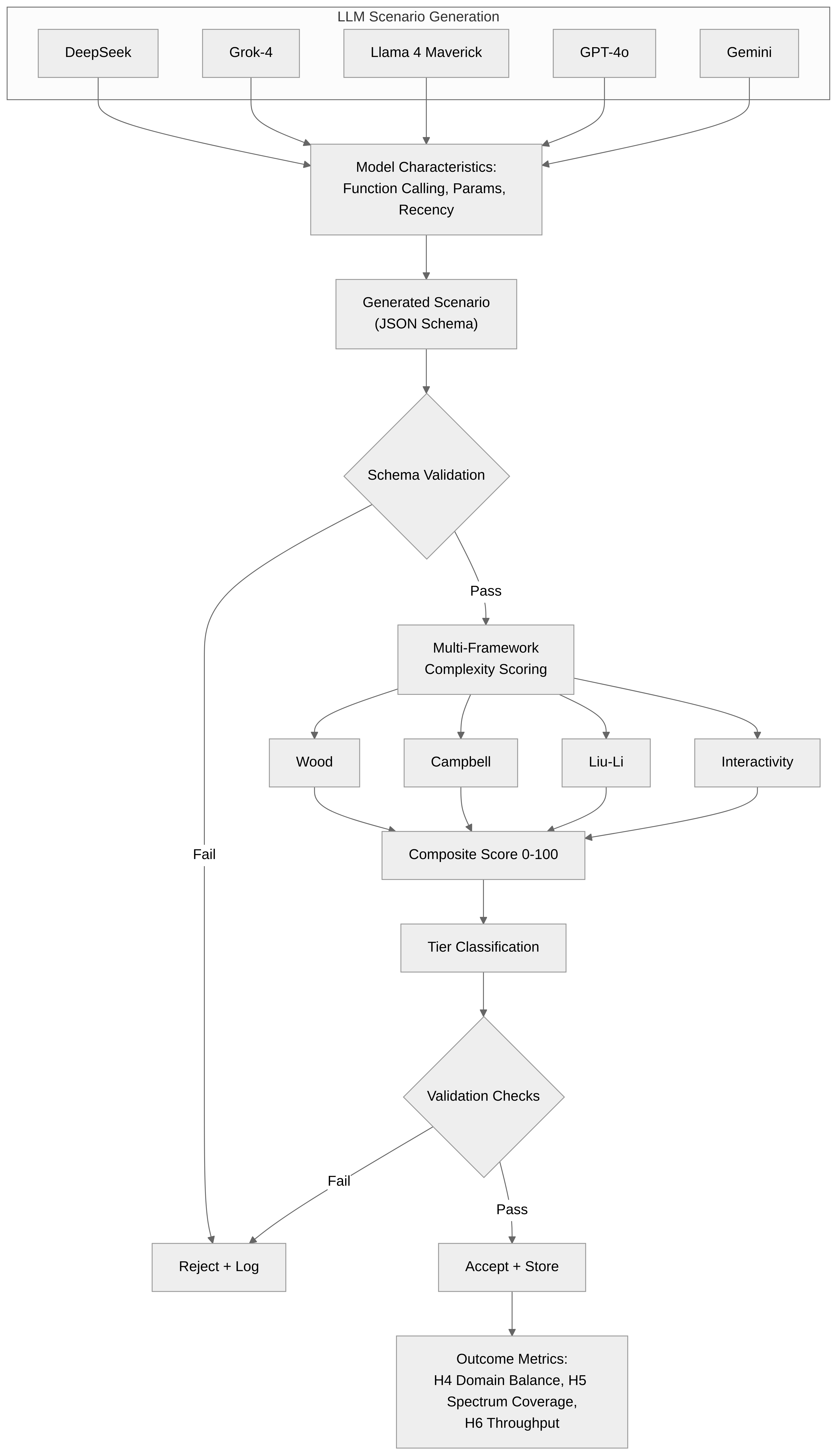}
  \caption{Two-stage measurement and analysis framework. Stage 1 establishes construct validity
    (H1a--H1e, H3); Stage 2 applies the validated instrument to compare model performance
    (H4--H6, RQ5--RQ6). Stage 2 is contingent on Stage 1.}
  \label{fig:two_stage}
\end{figure}

\begin{figure}[htbp]
  \centering
  \includegraphics[width=0.75\textwidth]{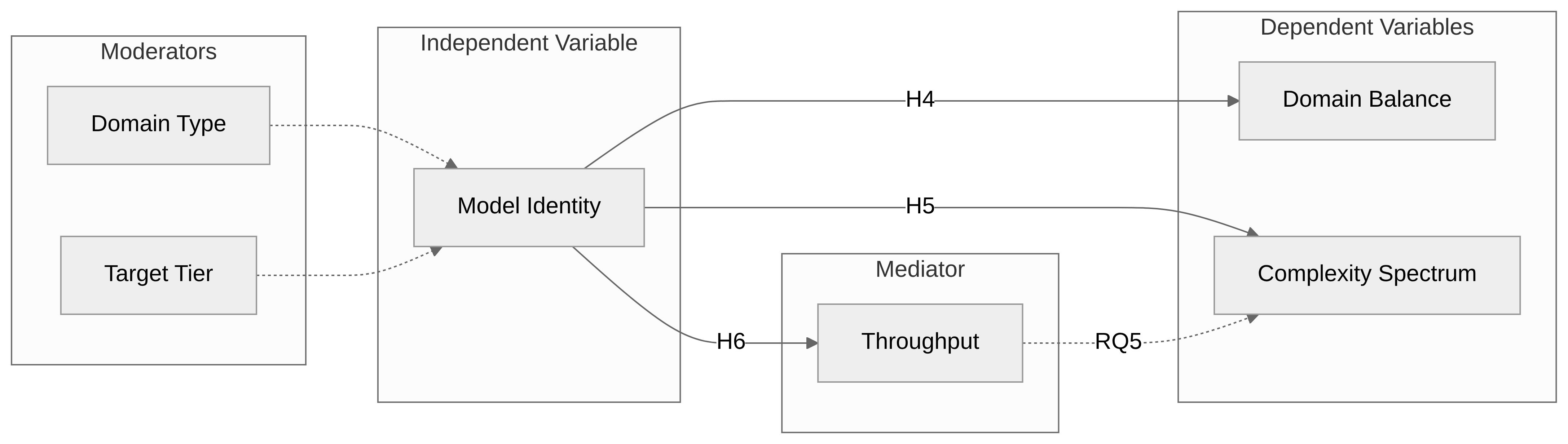}
  \caption{Analytical framework linking model identity, a mediating process, and performance
    outcomes. Model identity influences domain balance (H4), complexity-spectrum coverage (H5),
    and throughput (H6); throughput in turn relates to complexity coverage (RQ5, speed-quality
    trade-off). Domain type and target tier serve as moderators.}
  \label{fig:analytical}
\end{figure}

\begin{figure}[htbp]
  \centering
  \includegraphics[width=0.75\textwidth]{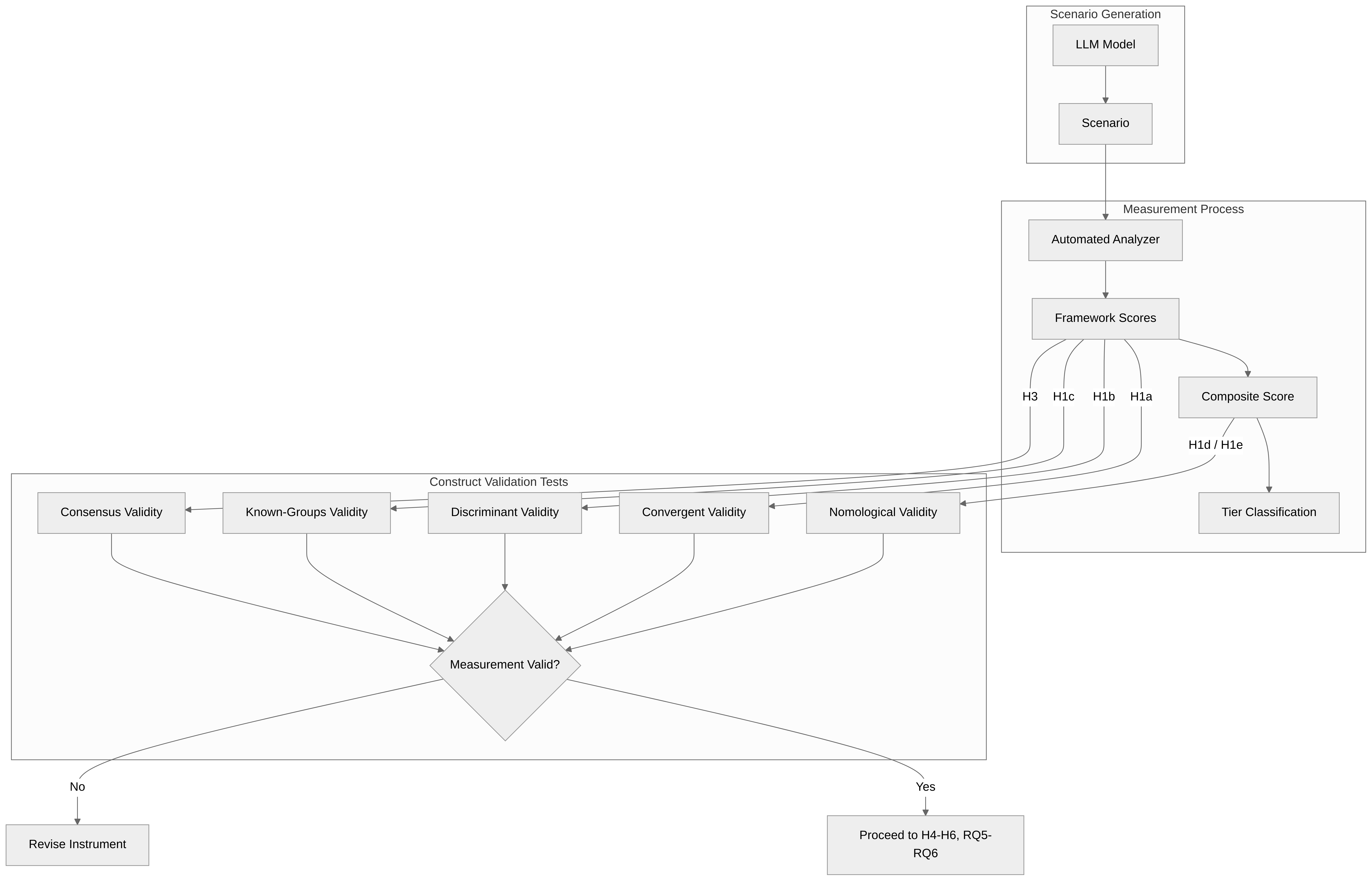}
  \caption{Measurement validation framework as prerequisite to model comparisons.
    Convergent validity, inter-LLM consensus, and known-groups validity must pass before
    model-level inferences are warranted.}
  \label{fig:measurement_validation}
\end{figure}

\begin{figure}[htbp]
  \centering
  \includegraphics[width=0.75\textwidth]{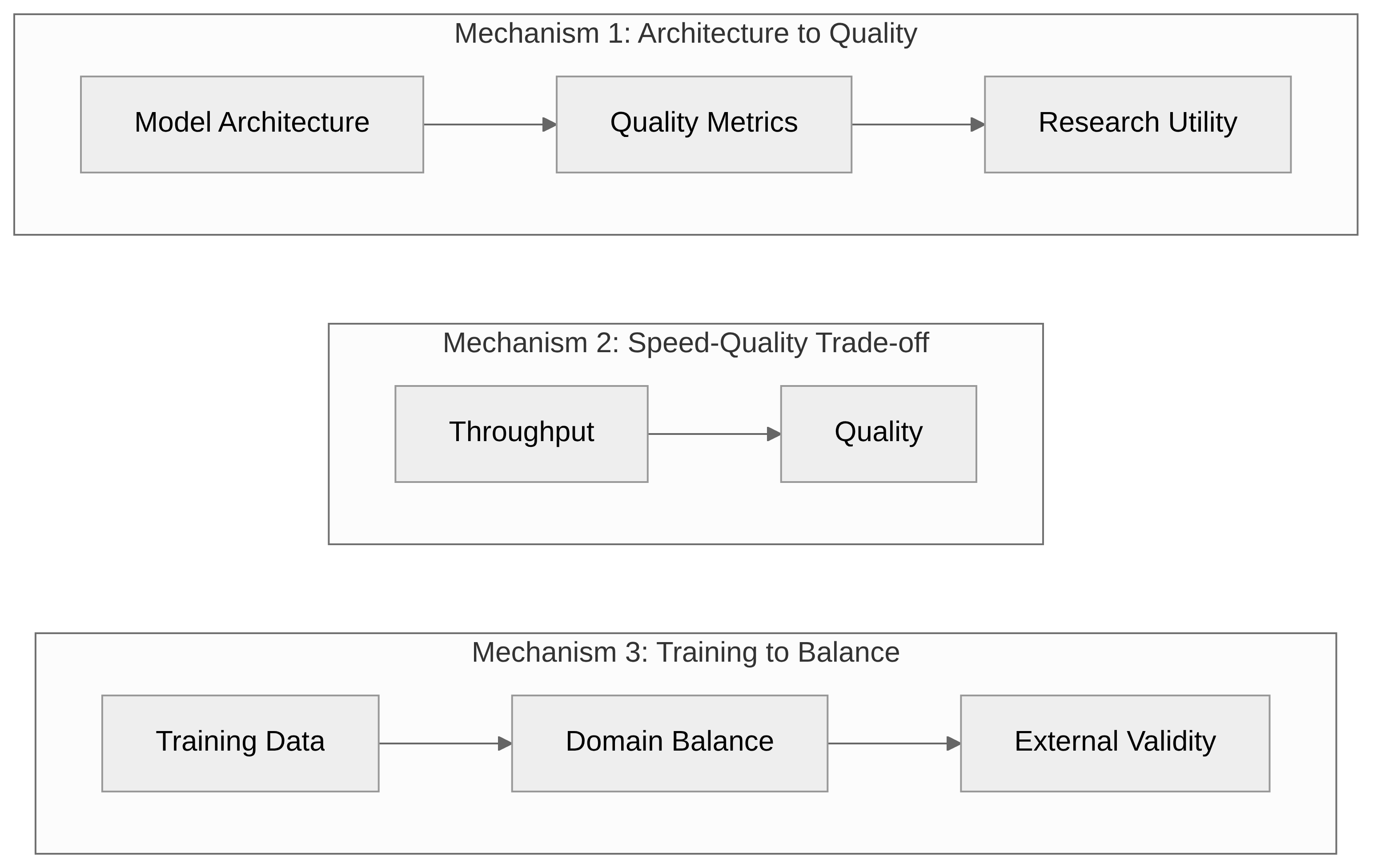}
  \caption{Proposed theoretical mechanisms linking model architectural properties to performance
    outcomes. Architectural constraints drive systematic trade-offs in speed, quality, and
    complexity coverage.}
  \label{fig:theoretical}
\end{figure}

\begin{figure}[htbp]
  \centering
  \includegraphics[width=0.75\textwidth]{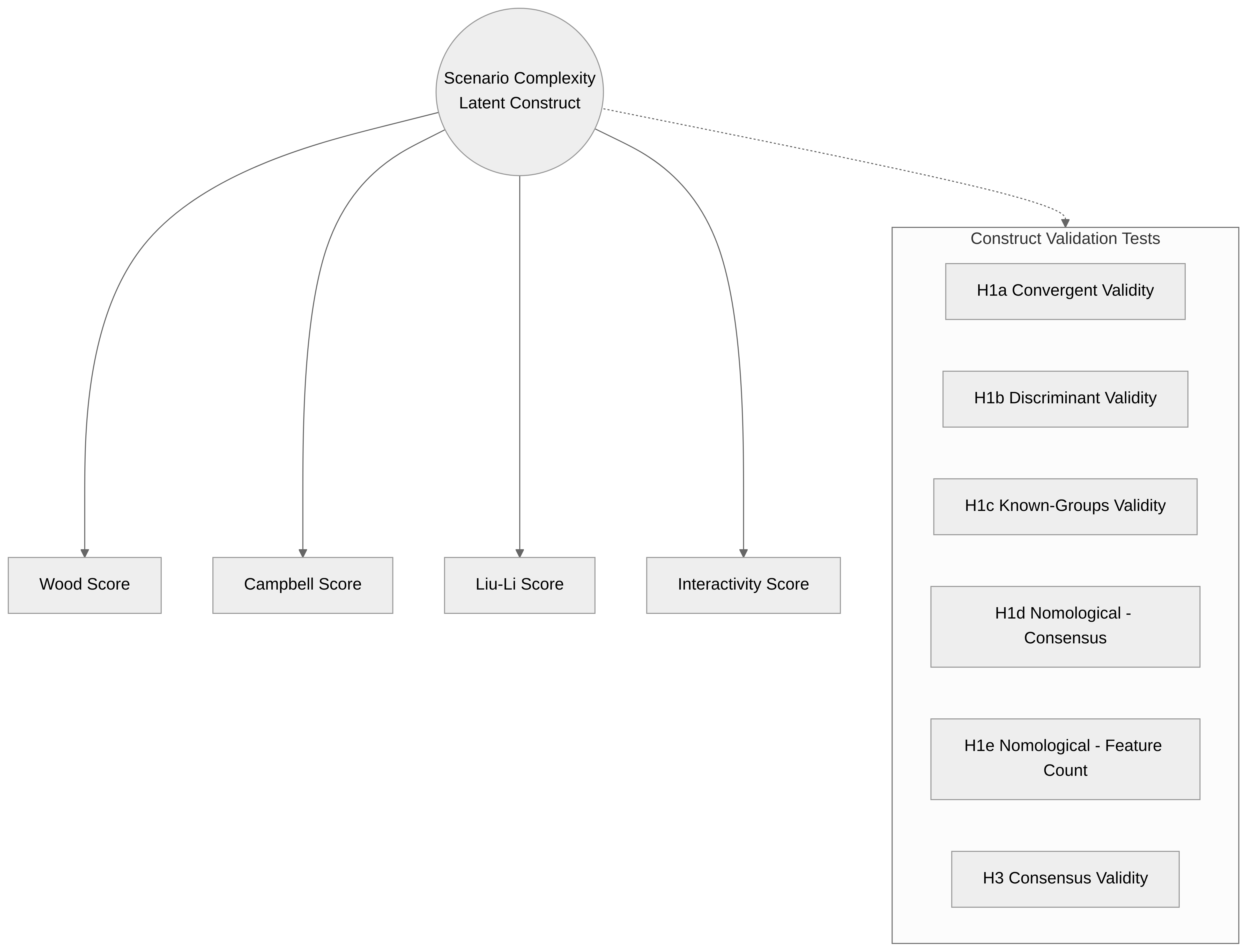}
  \caption{Proposed measurement model for the latent complexity construct. Wood, Campbell, and
    Liu \& Li load on a dominant structural complexity factor; Sweller's interactivity forms a
    secondary cognitive load dimension.}
  \label{fig:measurement_model}
\end{figure}

\clearpage
% -------------------------------------------------------
% 2. DATA COLLECTION AND METHODS
% -------------------------------------------------------
\FloatBarrier
\section{Data Collection and Methods}
\label{sec:methods}

\subsection{Scenario Generation}
\label{sec:generation}

Scenarios were generated by LLMs using a fixed prompt and a 14-field JSON schema across four
decision domains, Analytical, Planning, Communication, and Problem Solving, and three
complexity tiers: Simple, Moderate, and Complex. Temperature was held at 0 to minimize
stochastic variance. Scenarios failing schema validation were rejected and logged.

\begin{table}[htbp]
\centering
\caption{Scenario Generation Schema (14 Top-Level Fields)}
\label{tab:schema}
\small
\begin{tabular}{p{3.8cm}p{9cm}}
\toprule
\textbf{Field} & \textbf{Description} \\
\midrule
\texttt{task\_id}            & Unique task identifier. \\
\texttt{task\_title}         & Short scenario title. \\
\texttt{task\_description}   & Main decision task presented to the evaluator. \\
\texttt{business\_context}   & Background context framing the task. \\
\texttt{scenario\_type}      & \texttt{control} or \texttt{test}. \\
\texttt{scenario\_number}    & Repetition index within the generation run. \\
\texttt{requirements}        & List of required analytical outputs. \\
\texttt{success\_criteria}   & Observable success criteria for the scenario. \\
\texttt{complexity\_level}   & Intended tier: Simple, Moderate, or Complex. \\
\texttt{estimated\_duration} & Expected completion time. \\
\texttt{domain\_specific\_data} & Embedded data; includes \texttt{data\_elements},
  \texttt{calculations\_required}, \texttt{industry\_context}, \texttt{business\_function}. \\
\texttt{control\_expectations} & Blind reference; includes \texttt{expected\_calculations},
  \texttt{expected\_insights}, \texttt{expected\_trends}, \texttt{success\_thresholds}. \\
\texttt{cognitive\_assessment\_requirements} & Cognitive assessment generation metadata; includes
  \texttt{primary\_dimensions\_tested}, \texttt{cognitive\_complexity\_level},
  \texttt{memory\_integration\_opportunities}, \texttt{knowledge\_transfer\_elements},
  \texttt{expected\_tool\_usage\_patterns}. \\
\texttt{metadata}            & Provenance fields; includes \texttt{source\_agent},
  \texttt{experiment\_id}, \texttt{business\_domain}, \texttt{timestamp},
  \texttt{complexity\_justification}, \texttt{framework\_version},
  \texttt{cognitive\_assessment\_focus}. \\
\bottomrule
\end{tabular}
\end{table}

Because the schema contains a \texttt{cognitive\_assessment\_requirements} block, the resulting
scenario representation is not fully independent of the cognitive assessment instrument for
which the scenarios were designed. In the implementation used for this study, those fields
functioned as generation-side scaffolding and audit metadata, and portions were available to
the validation middleware when assembling scenario text for analyzer input. Accordingly, this
study validates a cognitive-assessment-compatible scenario-generation-and-scoring pipeline
rather than a strictly assessment-blind complexity instrument. This dependency should be treated
as a design limitation when interpreting construct purity.

\subsection{Complexity Scoring}
\label{sec:scoring}

Each scenario was scored using four frameworks:
\begin{itemize}[noitemsep]
  \item \textbf{Wood (1986):} Products, acts, and information cues, operationalized through
        component and coordinative complexity \citep{Baker2025,Stajkovic2025}.
  \item \textbf{Campbell (1988):} Multiple paths, conflicting information, and uncertain
        outcomes, quantified through decision-path enumeration \citep{Baker2025}.
  \item \textbf{Liu \& Li (2012):} Working memory load and problem-solving demands, assessed
        via information element analysis.
  \item \textbf{Sweller (1994):} Element interactivity, measured as the degree to which
        elements must be processed simultaneously \citep{Chen2023,Sweller2019}.
\end{itemize}

Each framework's analyzer maps a scenario's measured attributes to a framework sub-score, and
the four sub-scores are summed with equal unit weight to form the composite:

\begin{equation}
\text{CompositeScore} = \text{WoodScore} + \text{CampbellScore} + \text{LiuLiScore} + \text{InteractivityScore}
\label{eq:composite}
\end{equation}

The weighting scheme is the instrument's operationalization of the four frameworks, not a free
parameter estimated from data. The dimensions scored within each framework, and the ordinal
direction of their contributions, follow directly from the source theories: Wood's component,
coordinative, and dynamic complexity; Campbell's four sources of complexity (multiple paths,
multiple outcomes, conflicting interdependence, and uncertainty); Liu and Li's ten complexity
dimensions; and Sweller's element interactivity. The specific magnitudes were fixed during pilot
development to produce well-distributed composite scores across the three tiers; they are a
transparent, deterministic design choice rather than an empirically optimized parameterization,
and the weight-sensitivity analysis reported below shows that the instrument's tier
classifications do not depend on the particular values chosen. In summary, the Wood sub-score sums
distinct acts (capped at 10, weighted $0.5$), information cues per act (capped at 5, weighted
$1.0$), coordinative complexity (networked/interdependent/low $= 5/3/1$), and dynamic complexity
(high/low/none $= 4/2/0$); the Campbell sub-score accumulates graded contributions for multiple
paths, multiple outcomes, and conflicting interdependence (base weights $3$, $3$, and $4$, each
with a small count-based increment up to $+3$) plus an uncertainty term (high/bounded $= 5/3$);
the Liu \& Li sub-score sums ten dimensions (variety, ambiguity, relationships, unreliability,
and incongruity weighted $4$--$5$; novelty, variability, and time pressure as ordinal terms; and
size and action complexity as capped count terms); and the interactivity sub-score is
$\text{ratio} \times 10 + \text{depth} \times 0.6 + \text{edges} \times 0.2$. The complete weight
specification and the validation pass/fail rules are provided in Appendix~B.

Because the four frameworks operate on different raw scales, the additive composite is
empirically weighted toward the Liu \& Li framework, which alone accounts for roughly half of
the between-scenario variance in the composite (Liu \& Li $48\%$, Campbell $42\%$, Wood $8\%$,
interactivity $2\%$). We therefore standardize ($z$-score) the framework sub-scores for all
factor-analytic procedures, consistent with standard practice \citep{Campbell1959}. A
weight-sensitivity analysis (Section~\ref{sec:h1a}) confirms that this scale dependence does not
drive the instrument's classifications: across alternative weighting schemes, unit weighting,
scale-equalized ($z$-scored) weighting, and across-framework reweighting that removes the
Liu \& Li magnitude advantage, the composite rank-ordering is essentially unchanged
(Spearman $\rho \geq 0.98$) and tier separation remains large and stable
($\eta^2 \in [0.59, 0.62]$).

Predicted tiers are assigned from the composite using fixed cutoffs (Simple $0$--$20$, Moderate
$20$--$50$, Complex $50+$). These \emph{predicted} tiers are distinct from the \emph{intended}
(prompted) tier embedded at generation; the known-groups analysis (Section~\ref{sec:h1c}) uses
the intended tier as its grouping variable, not the composite-threshold tier.

The element-interactivity sub-score required a direction correction. The raw analyzer score
increases with the simultaneous-elements ratio in a direction that inflated low-complexity
scenarios; because higher element interactivity should denote greater intrinsic cognitive load,
we applied the transformation $\text{corrected} = 10.2 - \text{raw}$ (the additive constant
preserves the observed upper bound) so that the sub-score increases with cognitive load. All
factor analyses use the corrected interactivity score.

For H1a, exploratory factor analysis (EFA) used principal axis factoring with promax rotation
on standardized framework scores. Confirmatory factor analysis (CFA) was conducted in
\texttt{semopy}~2.3.11. The standard engine workflow supports a one-factor specification; the
two-factor model is a supplementary diagnostic specification. Because Interactivity was modeled
as a single-indicator latent in the two-factor specification, its error variance was fixed at
0.881, derived from its EFA communality.

\subsection{Validation Design}
\label{sec:design}

Measurement validation preceded model comparisons, and the validation procedures rely on two
distinct scoring systems that we distinguish explicitly. The complexity instrument itself
(H1a--H1e) is \emph{deterministic}: each scenario is scored by the fixed analytic procedure of
Section~\ref{sec:scoring}, which maps its attributes to the framework sub-scores and composite
with no stochastic component, so re-scoring the same scenario yields identical values. The
inter-rater analysis (H3) instead employs five independent LLM judges, each scoring scenarios
through its own model; agreement across these judges tests whether the complexity construct is
recovered consistently across different models rather than the reproducibility of the
deterministic scorer. Sampling for the validation subsets was stratified rather than
census-based: H3 used the five-judge consensus design, with the reported ICC(2,k) computed on
the 44-scenario fully crossed complete-case subset for which all five judges returned usable
ratings, and H1d/H1e used the consensus-scored subset available at the time of analysis
($N = 220$).

\subsection{Data Sources}
\label{sec:data}

The analyses draw on the complete set of scenarios produced during the study's generation
campaign. The primary validation corpus (\texttt{experiment\_scenarios}) comprises the
$N = 4{,}238$ scenarios that passed schema and complexity validation, generated by five models:
Grok-4 ($n = 1{,}195$), Llama~4 Maverick ($n = 1{,}081$), DeepSeek Chat~V3 ($n = 1{,}070$),
GPT-4o ($n = 890$), and a two-scenario GPT-5.2 pilot. This is the full validated output of the
campaign rather than a curated subset: Gemini~2.5 Pro produced no schema-valid scenarios and is
excluded, and Claude~3.7 Sonnet and \texttt{chatgpt-4o-latest} contributed only single
generation jobs and are excluded from model-level inference. Inter-LLM consensus scores
(\texttt{consensus\_validation\_results}, $N = 220$) support the agreement and nomological
analyses (H3, H1d/H1e). All datasets are available via Zenodo
(Section~\ref{sec:data_availability}).

\subsection{Statistical Analysis}
\label{sec:stats}

Pearson and Spearman correlations assessed convergent and nomological relationships. One-way
ANOVA with Tukey HSD evaluated known-groups validity. ICC(2,k) and Fleiss' $\kappa$ quantified
inter-LLM agreement. Partial correlations controlling for intended tier assessed the
relationship between complexity and text length (Section~\ref{sec:h1b}). Alpha was set at $.05$
unless otherwise specified.

For CFA reporting, the prespecified fit indices were CFI, RMSEA, and SRMR. Factor loadings are
interpreted alongside global fit; given the four-indicator identification constraint, the CFA
serves as a stress test rather than a primary confirmatory claim.

\textit{Unit of analysis.} H4--H5 use scenario-level counts aggregated by model; H6 uses
job-level throughput observations, preserving appropriate sample sizes despite a small number
of models.

\subsection{System Architecture}
\label{sec:architecture}

The system consists of a generation pipeline, a scoring and validation pipeline, a persistence
layer, and an analysis interface.

\begin{figure}[htbp]
  \centering
  \includegraphics[width=0.92\textwidth]{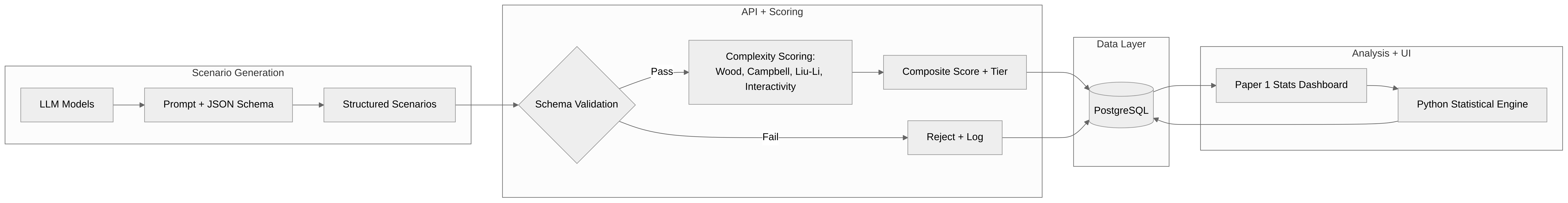}
  \caption{High-level system architecture. The generation pipeline (left) interfaces with LLM
    APIs; the scoring pipeline (center) applies four complexity frameworks; the persistence
    layer (right) stores scenarios and scores for downstream analysis.}
  \label{fig:architecture}
\end{figure}

% -------------------------------------------------------
% 3. RESULTS
% -------------------------------------------------------
\FloatBarrier
\section{Results}
\label{sec:results}

Table~\ref{tab:summary} summarizes the validation status for each hypothesis.

\begin{table}[htbp]
\centering
\caption{Summary of Hypothesis Tests and Research Questions}
\label{tab:summary}
\small
\resizebox{\textwidth}{!}{%
\begin{tabular}{p{1.2cm}p{3.5cm}p{6cm}p{2.5cm}}
\toprule
\textbf{Item} & \textbf{Test} & \textbf{Key Outcome} & \textbf{Status} \\
\midrule
H1a & EFA + AVE            & Dominant structural factor (62.5\% var; AVE~$=0.77$); weak interactivity & Supported \\
H1b & Correlations + ANOVA  & Domain passes; model moderate; word count strong               & Mixed           \\
H1c & One-way ANOVA + Tukey & Large tier separation ($\eta^{2} = 0.587$)                      & Full Support    \\
H3  & ICC(2,k) + $\kappa$   & ICC~$= 0.997$; $\kappa = 0.971$                                & Full Support    \\
H1d & Pearson $r$           & $r = -0.437$ (complexity $\to$ lower consensus)                & Full Support    \\
H1e & Pearson + Spearman    & $r = 0.644$; $\rho = -0.371$                                   & Full Support    \\
H4  & Chi-square GOF        & DeepSeek Chat V3/Grok/Llama balanced; GPT-4o biased                    & Full Support    \\
H5  & Chi-square indep.     & $\chi^{2} = 21.56$, $p = 0.006$, $V = 0.050$                  & Full Support    \\
H6  & ANOVA + Tukey HSD     & $F = 23.69$, $p = 8.52\times10^{-5}$, $\eta^{2} = 0.913$      & Full Support    \\
RQ5 & Pearson $r$ (speed--quality) & $r = -0.967$ ($n = 5$); near zero excluding Llama        & Suggestive      \\
RQ6 & Chi-square (model $\times$ failure) & $\chi^{2} = 249.16$, $V = 0.709$ (Llama vs.\ Grok) & Supported    \\
\bottomrule
\end{tabular}}
\end{table}

\subsection{H1a: Convergent Validity}
\label{sec:h1a}

\begin{table}[htbp]
\centering
\caption{Framework Intercorrelations After Interactivity Inversion ($n = 4{,}238$)}
\label{tab:intercorr}
\begin{tabular}{lcccc}
\toprule
\textbf{Framework} & \textbf{Wood} & \textbf{Campbell} & \textbf{Liu-Li} & \textbf{Interactivity} \\
\midrule
Wood          & 1.00 & 0.64 & 0.83 & 0.30 \\
Campbell      & 0.64 & 1.00 & 0.81 & 0.30 \\
Liu-Li        & 0.83 & 0.81 & 1.00 & 0.31 \\
Interactivity & 0.30 & 0.30 & 0.31 & 1.00 \\
\bottomrule
\end{tabular}
\end{table}

\begin{table}[htbp]
\centering
\caption{Factor Loadings: EFA and CFA (Standardized)}
\label{tab:efa}
\small
\begin{tabular}{lcccccc}
\toprule
 & \multicolumn{3}{c}{\textbf{EFA}} & \multicolumn{3}{c}{\textbf{CFA (two-factor)}} \\
\cmidrule(lr){2-4}\cmidrule(lr){5-7}
\textbf{Variable} & \textbf{F1 $\lambda$} & \textbf{F2 $\lambda$} & \textbf{$h^{2}$} & \textbf{Factor} & \textbf{$\beta$} & \textbf{$h^{2}$} \\
\midrule
scoreWood          & 0.87 &  0.28   & 0.84 & Structural     & 0.900 & 0.810 \\
scoreCampbell      & 0.84 & $-0.32$ & 0.81 & Structural     & 0.888 & 0.788 \\
scoreLiuLi         & 0.96 & $-0.01$ & 0.92 & Structural     & 0.962 & 0.925 \\
scoreInteractivity & 0.34 & $-0.02$ & 0.24 & Cognitive load & 0.345 & 0.119 \\
\bottomrule
\end{tabular}
\tablenote{EFA: principal axis factoring; $h^{2}$ is communality. CFA standardized loadings
($\beta$) are all $p < .001$ (SE $\le 0.014$). The two solutions are reported together; the
convergent-validity argument rests on the EFA (see text).}
\end{table}

\noindent\textbf{Model Fit.} The exploratory solution (Table~\ref{tab:efa}; principal axis
factoring, overall $\mathrm{KMO} = 0.69$, per-framework $0.62$--$0.93$; Bartlett's
$\chi^{2} = 9{,}972$, $p < .001$) accounts for
$67.0\%$ of total variance: a dominant structural factor explaining $62.5\%$ and a weak secondary
cognitive-load factor explaining $4.5\%$. The confirmatory specifications fit poorly on global
indices, one-factor CFA: CFI~$= .913$, RMSEA~$= .321$, 90\%~CI~[.320, .321], SRMR~$= .087$;
two-factor CFA: CFI~$= .858$, RMSEA~$= .579$, 90\%~CI~[.578, .579], SRMR~$= .175$.

These fit indices reflect the structural constraints of a four-indicator model rather than
substantive misspecification. RMSEA is known to be inflated and unstable in models with very low
degrees of freedom, precisely the regime here, where four indicators (one of them a
single-indicator latent with fixed error variance) leave little identification headroom
\citep{KennyKaniskanMcCoach2015}. The two-factor specification fits \emph{worse} than the
one-factor not because two-dimensionality is wrong but because adding a single-indicator second
latent to a near-saturated model degrades global fit; the inflated inter-factor estimate
($\varphi = .963$) is a mathematical consequence of that single-indicator specification, not a
finding of construct overlap. We therefore treat the CFA as a diagnostic stress test rather than
confirmatory evidence and base the convergent-validity argument on the EFA.

On that basis convergent validity is supported. The three structural indicators load strongly on
the dominant factor (Wood $.87$, Campbell $.84$, Liu \& Li $.96$), yielding an average variance
extracted of $\mathrm{AVE} = 0.77$ ($\sqrt{\mathrm{AVE}} = 0.88$), well above the $0.50$ threshold
for convergent validity \citep{FornellLarcker1981}. The interactivity indicator loads weakly
($\lambda = .34$); this is primarily a property of the data rather than a measurement failure.
Element interactivity has very low between-scenario variance in this corpus, it contributes only
about $2\%$ of composite-score variance, so it cannot covary strongly with the factor that
discriminates scenarios, consistent with its theoretical status as a distinct cognitive-load
dimension (Section~\ref{sec:framework}). We report the variance explained by factor rather than as
a single pooled figure precisely because two of the four indicators (Wood and interactivity)
carry little between-scenario variance, which structurally bounds the total.

\textit{Conclusion:} Convergent validity is supported, a clear dominant structural factor
(AVE $= 0.77$) with a theoretically distinct, weakly varying interactivity dimension. At the
raw-component level, the interactivity score is positively (if modestly) aligned with the other
components after the inversion correction (pairwise $r = 0.20$--$0.32$), consistent with its
treatment as a distinct cognitive-load dimension.

\subsection{H1c: Known-Groups Validity}
\label{sec:h1c}

One-way ANOVA revealed large, ordered differences across intended tiers
(Simple $<$ Moderate $<$ Complex), with a very large effect size ($\eta^{2} = 0.587$); all
pairwise comparisons were significant ($p < .001$).

The grouping variable is the \emph{intended} tier embedded in the generation prompt, not the
composite-threshold (predicted) tier (Section~\ref{sec:scoring}). The analysis therefore tests
whether the instrument recovers the complexity level the generator was instructed to produce and
is free of the circularity that would arise from grouping scenarios by the very threshold applied
to their own composite scores. This establishes \emph{internal} criterion validity: the
instrument reliably reproduces an externally specified design factor. It does not, by itself,
establish \emph{external} criterion validity against human outcomes such as decision latency,
error rates, or expert difficulty ratings; that comparison is a downstream study
(Section~\ref{sec:future}).
\begin{table}[htbp]
\centering
\caption{H1c Known-Groups Validity: Tier Descriptives ($n = 4{,}238$)}
\label{tab:kgv}
\begin{tabular}{lcccc}
\toprule
\textbf{Tier} & \textbf{N} & \textbf{M} & \textbf{SD} & \textbf{95\% CI} \\
\midrule
Simple   & 1,465 & 16.48 &  4.21 & [16.26, 16.70] \\
Moderate & 1,411 & 30.00 &  8.38 & [29.56, 30.44] \\
Complex  & 1,362 & 43.49 & 13.19 & [42.79, 44.19] \\
\bottomrule
\end{tabular}
\tablenote{One-way ANOVA: $F(2,\,4235) = 3{,}013.69$, $p < .001$, $\eta^{2} = 0.587$. All three
Tukey HSD pairwise contrasts are significant ($p_{\text{adj}} < .001$); mean differences are
Complex--Moderate $= 13.49$, Moderate--Simple $= 13.52$, Complex--Simple $= 27.01$.}
\end{table}

\subsection{H1b: Discriminant Validity}
\label{sec:h1b}

Discriminant validity has two facets here: whether the two latent constructs are distinct from
one another, and whether the composite is separable from potential confounds. On the first, the
Fornell-Larcker criterion is satisfied: the square root of the average variance extracted for the
structural-complexity construct ($\sqrt{\mathrm{AVE}} = 0.88$) and for the cognitive-load
construct ($\sqrt{\mathrm{AVE}} = 0.49$) both exceed their inter-construct correlation
($r \approx 0.31$), indicating that the structural and cognitive-load dimensions are empirically
distinct \citep{FornellLarcker1981}.

Among potential confounds, domain effects were negligible ($\eta^{2} = 0.006$, all frameworks;
Table~\ref{tab:h1b}), supporting domain generality. Model identity showed a moderate
effect ($\eta^{2} = 0.193$); as elaborated in Section~\ref{sec:discuss_disc}, this is not in
tension with the near-perfect inter-judge agreement (ICC~$= 0.997$), it reflects that different
models \emph{generate} scenarios of differing complexity, whereas the ICC reflects that judges
\emph{score the same scenarios} consistently.

A stronger association was observed for text length: complexity correlated with word count at
$r = 0.91$. Because the intended tiers differ in both complexity and length by design, we
examined whether this association reflects only the tiered structure by computing the partial
correlation controlling for intended tier, together with the within-tier correlations
(Table~\ref{tab:wc_partial}). The relationship
persists after removing all tier-level variation (partial $r = 0.86$) and is strongest within
the Complex tier ($r = 0.88$). We therefore report length dependence as a scope boundary of the
instrument rather than claiming construct-pure separation from length; in this corpus the
dependence is concentrated in the Liu \& Li component (Section~\ref{sec:discuss_disc}).

\begin{table}[htbp]
\centering
\caption{Complexity--Word-Count Association Controlling for Intended Tier ($n = 4{,}238$)}
\label{tab:wc_partial}
\begin{tabular}{lcl}
\toprule
\textbf{Analysis} & \textbf{$r$} & \textbf{Interpretation} \\
\midrule
Zero-order (overall)          & 0.91 & Strong \\
Partial, controlling for tier & 0.86 & Persists after removing tier variation \\
Within Simple tier            & 0.73 & \\
Within Moderate tier          & 0.83 & \\
Within Complex tier           & 0.88 & Strongest within-tier association \\
\bottomrule
\end{tabular}
\end{table}
\begin{table}[htbp]
\centering
\caption{H1b Discriminant Validity Summary}
\label{tab:h1b}
\begin{tabular}{lllll}
\toprule
\textbf{Analysis} & \textbf{Statistic} & \textbf{Value} & \textbf{Criterion} & \textbf{Status} \\
\midrule
Complexity $\times$ Word Count (zero-order) & Pearson $r$ & 0.91  & $r < 0.30$        & Bounded \\
Complexity $\times$ Word Count (partial $\mid$ tier) & Partial $r$ & 0.86 & n/a        & Persists \\
Construct discriminant (Fornell-Larcker) & $\sqrt{\mathrm{AVE}}$ vs $r$ & 0.88, 0.49 $>$ 0.31 & $\sqrt{\mathrm{AVE}} > r$ & Pass \\
Complexity $\times$ Model ID   & $\eta^{2}$  & 0.193 & $\eta^{2} < 0.10$ & Exceeds \\
Complexity $\times$ Domain     & $\eta^{2}$  & 0.006 & $\eta^{2} < 0.15$ & Pass    \\
\bottomrule
\end{tabular}
\end{table}

Descriptively, mean composite complexity varied little across the four task domains
($28.6$--$31.5$) but more across models, ranging from $20.2$ (Llama~4 Maverick) to $35.3$
(Grok-4); this is consistent with the negligible domain effect and the moderate model effect
reported above. The two-scenario GPT-5.2 pilot ($n = 2$, $M = 27.8$) is excluded from all
comparisons.

\textit{Conclusion:} Mixed. Confound with word count is expected and defensible given the frameworks'
operational definitions of complexity \citep{Campbell1988,Wood1986}.

\subsection{H3: Inter-LLM Consensus Validity}
\label{sec:h3}

Inter-LLM agreement was near-perfect on continuous scores (ICC~$= 0.997$) and categorical tiers
(Fleiss' $\kappa = 0.971$), far exceeding thresholds for excellent reliability
\citep{Cicchetti1994}. The five-judge panel was drawn from five independent
organizations and architectural lineages, GPT-4 Turbo (OpenAI), DeepSeek Chat~V3 (DeepSeek),
Claude Sonnet~4.5 (Anthropic), Mistral Large (Mistral), and Llama~4 Maverick (Meta), so that
agreement reflects convergence across distinct training data and alignment regimes rather than
shared within-family bias. Agreement among such heterogeneous judges indicates that the
complexity signal is reproducible across scorers; as with any consensus measure, however, it
establishes consistency rather than correctness against an external human standard, which we
identify as a complementary validation step (Section~\ref{sec:future}). The reported coefficients
were estimated on the 44-scenario fully crossed complete-case subset for which all five judges
returned usable ratings.

\begin{table}[htbp]
\centering
\caption{H3 Consensus Agreement}
\label{tab:consensus}
\begin{tabular}{lll}
\toprule
\textbf{Metric} & \textbf{Value} & \textbf{Interpretation} \\
\midrule
ICC(2,k)              & 0.9971   & Excellent         \\
Fleiss' $\kappa$      & 0.9714   & Almost perfect    \\
$n_{\text{subjects}}$ & 44       & n/a               \\
$n_{\text{raters}}$   & 5        & Independent LLMs  \\
$p$                   & $< .001$ & Significant       \\
\bottomrule
\end{tabular}
\end{table}

\subsection{H1d/H1e: Nomological Validity}
\label{sec:nomological}

\begin{table}[htbp]
\centering
\caption{H1d/H1e Nomological Validity Correlations ($N = 220$)}
\label{tab:nomological}
\begin{tabular}{lllll}
\toprule
\textbf{Relationship} & \textbf{Statistic} & \textbf{Value} & \textbf{p} & \textbf{Interpretation} \\
\midrule
Complexity $\times$ Consensus & Pearson $r$     & $-0.437$ & $1.15\times10^{-11}$ & Moderate negative \\
Complexity $\times$ Features  & Pearson $r$     &  0.644   & $3.68\times10^{-27}$ & Strong positive   \\
Complexity $\times$ Consensus & Spearman $\rho$ & $-0.371$ & $1.41\times10^{-8}$  & Robust negative   \\
\bottomrule
\end{tabular}
\tablenote{Subset drawn from scenarios with completed consensus-evaluation records at analysis
execution time.}
\end{table}

\textit{Conclusion:} Full support; complexity behaves as predicted in the nomological network.

\subsection{Scoring Reproducibility}
\label{sec:testretest}

We do not report a test-retest reliability coefficient. The complexity scorer is deterministic
(Section~\ref{sec:design}): for a fixed scenario it returns identical framework sub-scores and
composite on every run, so reproducibility is exact by construction, and a temporal-stability
coefficient over a re-scoring interval is not a meaningful quantity for this instrument. Temporal stability is instead a property of the LLM-judge layer; characterizing
judge drift over an interval is left to future work (Section~\ref{sec:future}). The reliability
evidence relevant to scoring is therefore the inter-judge agreement reported in H3
(Section~\ref{sec:h3}; ICC~$= 0.997$ across five independent model families), which tests whether
the construct is recovered consistently across different scorers.

\subsection{H4: Domain Balance}
\label{sec:h4}

Chi-square goodness-of-fit tests assessed whether each model produced balanced domain
coverage. DeepSeek Chat V3, Grok-4, and Llama~4 Maverick showed near-perfect balance; GPT-4o
was severely biased ($p < .001$).

\noindent GPT-5.2 had only two scenarios and was excluded from inference due to insufficient $N$.
\begin{table}[htbp]
\centering
\caption{H4 Domain Balance Chi-Square}
\label{tab:domain_balance}
\begin{tabular}{lllll}
\toprule
\textbf{Model} & \textbf{$\chi^{2}$} & \textbf{p} & \textbf{N} & \textbf{Interpretation} \\
\midrule
DeepSeek Chat V3 &   0.019 & 0.999                 & 1,070 & Balanced \\
Grok-4           &   0.023 & 0.999                 & 1,195 & Balanced \\
Llama 4 Maverick &   0.935 & 0.817                 & 1,081 & Balanced \\
GPT-4o           & 584.724 & $2.07\times10^{-126}$ &   890 & Biased   \\
\bottomrule
\end{tabular}
\tablenote{GPT-5.2 ($n = 2$) excluded from inference.}
\end{table}

\subsection{H5: Complexity Spectrum Coverage}
\label{sec:h5}

Chi-square independence test confirmed significant tier distribution differences across
models ($\chi^{2} = 21.56$, $p = 0.006$), though the effect was small (Cramer's $V = 0.050$).
DeepSeek Chat V3 and Grok-4 showed near-uniform tier coverage, while GPT-4o and Llama~4
showed modest deviations.
\begin{table}[htbp]
\centering
\caption{H5 Complexity Spectrum Coverage}
\label{tab:spectrum}
\begin{tabular}{ll}
\toprule
\textbf{Statistic} & \textbf{Value} \\
\midrule
$\chi^{2}$   & 21.56  \\
df           & 8      \\
$p$          & 0.0058 \\
Cramer's $V$ & 0.050  \\
$N$          & 4,238  \\
\bottomrule
\end{tabular}
\end{table}

\subsection{H6: Throughput Variation}
\label{sec:h6}

\begin{table}[htbp]
\centering
\caption{H6 Throughput by Model (Scenarios/Min, Job-Level)}
\label{tab:throughput_means}
\begin{tabular}{lccc}
\toprule
\textbf{Model} & \textbf{n (jobs)} & \textbf{M} & \textbf{SD} \\
\midrule
DeepSeek Chat V3 & 4 &   25.25 & 24.71 \\
Grok-4           & 2 &   31.32 &  1.55 \\
GPT-4o           & 4 &   14.40 & 12.46 \\
Llama 4 Maverick & 2 &  134.29 &  2.47 \\
Gemini 2.5 Pro   & 2 &    0.00 &  0.00 \\
\bottomrule
\end{tabular}
\tablenote{One-way ANOVA on job-level rates: $F = 23.69$, $p = 8.52\times10^{-5}$,
$\eta^{2} = 0.913$ ($k = 5$). Tukey HSD: every significant contrast involves Llama~4 Maverick
(mean differences $103$--$134$ scenarios/min, all $p_{\text{adj}} < .001$); no other pair differs
significantly. Values are the mean of per-job throughput rates, distinct from the lifetime
aggregate throughput (total stored $\div$ total time) reported in Table~\ref{tab:model_perf}; the
two differ when a model's jobs vary in size or duration. Gemini~2.5 Pro's $0.00$ reflects complete
generation failure (no scenarios produced) rather than a measured rate, and is interpreted as a
failure condition. Models with only one job were excluded from inference:
\texttt{chatgpt-4o-latest} (a ChatGPT-routing alias for GPT-4o),
\texttt{claude-3-7-sonnet-20250219} (Claude 3.7 Sonnet), and GPT-5.2 (accessed via the
\texttt{openai/gpt-5.2} routing path through an OpenAI-compatible aggregator endpoint).}
\end{table}

\subsection{Dropped Analyses: H7 and RQ4}
\label{sec:dropped}

\textbf{H7 (Schema Compliance)} is dropped due to insufficient data: all models used function
calling, leaving no non-function-calling comparison group. This entry is retained to document the
planned analysis and data limitation.

\textbf{RQ4 (Predictive Model Characteristics)} is dropped due to insufficient metadata (parameter
count and training recency unavailable) and insufficient predictor variation to support logistic
regression.

\subsection{RQ5: Speed-Quality Trade-off}
\label{sec:rq5}

Across the five models, generation throughput was negatively correlated with schema pass rate
($r = -0.967$, $p = 0.007$, $n = 5$), where pass rate denotes the proportion of generated
scenarios satisfying all schema and validation criteria. This aggregate relationship is sensitive
to individual observations. A leave-one-out analysis indicates that it depends substantially on
Llama~4 Maverick, the one high-throughput model: excluding it reduces the correlation to near zero
($r = -0.03$), whereas excluding any other model leaves it largely unchanged ($|r| \geq 0.96$). A
rank-based coefficient computed without the two-scenario GPT-5.2 pilot is correspondingly modest
($\rho = -0.20$). The relationship is therefore best regarded as suggestive rather than
established: it is consistent with a speed-quality trade-off but rests on five model-level
aggregates and is influenced by a single high-leverage observation. A larger panel of models is
required to confirm the pattern (Section~\ref{sec:future}).

\subsection{RQ6: Failure Pattern Analysis}
\label{sec:rq6}

Category-level rejection records were retained for two models (Llama~4 Maverick and Grok-4). The
association between model and failure category is strong: $\chi^{2}(5) = 249.16$,
$p = 8.34\times10^{-52}$, Cramer's $V = 0.709$ ($N = 496$). The two models' failure profiles
differ sharply.

\begin{table}[htbp]
\centering
\caption{RQ6 Failure Mode Frequencies (Top Categories)}
\label{tab:failure_freq}
\begin{tabular}{llcc}
\toprule
\textbf{Model} & \textbf{Category} & \textbf{Count} & \textbf{\%} \\
\midrule
Llama 4 Maverick & scoreTooLow             & 98 & 19.76 \\
Llama 4 Maverick & tierMismatch            & 98 & 19.76 \\
Llama 4 Maverick & structuralIssues        & 98 & 19.76 \\
Llama 4 Maverick & missingComplexity       & 98 & 19.76 \\
Llama 4 Maverick & insufficientInformation & 98 & 19.76 \\
Grok-4           & lowConfidence           &  3 &  0.60 \\
Grok-4           & structuralIssues        &  2 &  0.40 \\
Grok-4           & tierMismatch            &  1 &  0.20 \\
\bottomrule
\end{tabular}
\end{table}

The rejection records show a model-specific pattern. Llama~4 Maverick's failures are concentrated
at the Complex tier and distributed evenly across five categories (scoreTooLow, tierMismatch,
structuralIssues, missingComplexity, and insufficientInformation; $n = 98$ each). This pattern is
consistent with a fast but structurally shallow generation process: Llama produced the highest
throughput (Section~\ref{sec:h6}), yet its complex-tier scenarios frequently fell short of the
complexity threshold and were rejected, which accounts for its lower proportion of complex
scenarios relative to the balanced models. Grok-4's rejections were sparse and predominantly
confidence-related (the \texttt{lowConfidence} category is over-represented at a standardized
residual of $15.56$, the largest in the table), consistent with its balanced, high-compliance
profile.

Two caveats apply. First, per-category rejection logs were retained only for
Llama~4 Maverick and Grok-4; the validation pipeline discarded failing scenarios from the other
models without category-level records, so their failures are characterized through the
distributional analyses rather than rejection categories. GPT-4o's quality limitation manifests
as severe domain imbalance ($\chi^{2} = 584.7$; Section~\ref{sec:h4}) and incomplete generation
rather than complex-tier rejection, and Gemini~2.5 Pro produced no schema-valid scenarios at all.
Second, the chi-square association ($N = 496$) therefore characterizes the Llama-versus-Grok
contrast specifically rather than all five models.

\subsection{Synthesized Model Performance}
\label{sec:model_summary}

\begin{table}[htbp]
\centering
\caption{Model Performance Snapshot}
\label{tab:model_perf}
\small
\resizebox{\textwidth}{!}{%
\begin{tabular}{lcccccc}
\toprule
\textbf{Model} & \textbf{N} & \textbf{Pass \%} & \textbf{Schema \%} &
\textbf{Throughput} & \textbf{Dom.\ Bal.} & \textbf{Complex \%} \\
\midrule
DeepSeek Chat V3 & 1,070 & 99.07 & 99.07 &  21.63 & 99.58 & 33.64 \\
Grok-4           & 1,195 & 99.58 & 99.58 &  31.28 & 99.57 & 33.14 \\
GPT-4o           &   890 & 98.67 & 98.67 &   4.77 & 18.94 & 34.61 \\
Llama 4 Maverick & 1,081 & 90.08 & 90.08 & 134.27 & 97.06 & 27.57 \\
\bottomrule
\end{tabular}}
\tablenote{Throughput is the lifetime aggregate rate (total stored $\div$ total generation time)
in scenarios/min, distinct from the per-job mean in Table~\ref{tab:throughput_means}. Pass~\% and
Schema~\% both denote schema compliance among stored scenarios and therefore coincide; this is
distinct from the target-completion rate (stored $\div$ attempted) referenced elsewhere. The
two-scenario GPT-5.2 pilot ($n = 2$) is excluded from all comparisons.}
\end{table}

\begin{table}[htbp]
\centering
\caption{Open Tests and Data Gaps}
\label{tab:gaps}
\begin{tabular}{p{4cm}p{2cm}p{7cm}}
\toprule
\textbf{Item} & \textbf{Status} & \textbf{Reason} \\
\midrule
H7 Schema Compliance          & Dropped & No non-function-calling comparison group \\
RQ4 Predictive Characteristics & Dropped & Insufficient metadata and predictor variance \\
RQ6 Coverage                  & Limited & Rejection records for Llama 4 and Grok-4 only \\
\bottomrule
\end{tabular}
\end{table}

% -------------------------------------------------------
% 4. DISCUSSION
% -------------------------------------------------------
\FloatBarrier
\section{Discussion}
\label{sec:discussion}

\subsection{Measurement Validation: Core Findings}
\label{sec:discuss_mv}

The measurement system passed its critical validation thresholds, establishing the instrument as
scientifically sound. \textbf{Inter-LLM consensus} was near-perfect (ICC~$= 0.997$,
95\%~CI~[0.995, 0.998]) across five independent model families, indicating that the complexity
construct is recovered consistently regardless of which model performs the scoring
\citep{Cicchetti1994}. Because the scoring instrument itself is deterministic, scoring
reproducibility is exact by construction (Section~\ref{sec:testretest}); the consensus result is
thus the substantive reliability evidence, testing agreement across scorers rather than mere
re-computation.

\textbf{Known-groups validity} showed the instrument discriminates complexity tiers with very large
effects: $F(2, 4235) = 3{,}013.69$, $p < .001$, $\eta^{2} = 0.587$. This means 58.7\% of
complexity variance is explained by intended tier membership, far exceeding the $\eta^{2} = 0.14$
threshold for large effects \citep{Cohen1988}. Simple scenarios averaged 16.5, Moderate 30.0,
and Complex 43.5, showing clear ordered separation consistent with theoretical expectations.

The \textbf{factor structure} revealed a dominant complexity construct explaining 62.5\% of total
variance (67.0\% for the two-factor solution; Section~\ref{sec:h1a}), with three frameworks loading
strongly ($\lambda = 0.87$--$0.96$: Wood, Campbell, Liu \& Li). Interactivity formed a weaker secondary dimension ($\lambda = 0.34$), theoretically
defensible given Sweller's focus on working memory constraints as a mechanism orthogonal to
structural complexity \citep{Chen2023,Paas2020,Sweller2020}. The intercorrelations
(Table~\ref{tab:intercorr}) support this: Wood, Campbell, and Liu \& Li correlate at $r = 0.64$--$0.83$
with each other but only $r = 0.30$--$0.31$ with interactivity.

\textbf{Nomological validity} confirmed the instrument behaves as complexity theory predicts.
More complex scenarios yielded significantly lower inter-judge consensus ($r = -0.437$,
$p = 1.15\times10^{-11}$) and activated significantly more framework features ($r = 0.644$,
$p = 3.68\times10^{-27}$). The two indicators operate in opposite directions, negative for
consensus, positive for feature count, and this divergence is theoretically expected rather than
contradictory. A single construct driving two indicators in opposite but theoretically predicted
directions is the hallmark of a well-specified nomological network \citep{Cronbach1955}.

\subsection{Discriminant Validity: The Word Count Problem}
\label{sec:discuss_disc}

Domain independence was excellent ($\eta^{2} = 0.006$), confirming complexity generalizes across
decision contexts without domain-specific recalibration. Model identity accounted for a moderate
share of score variance ($\eta^{2} = 0.193$), which is not in tension with the near-perfect
inter-judge agreement (ICC~$= 0.997$): the two statistics measure different things. The
$\eta^{2} = 0.193$ reflects that different models \emph{generate} scenarios of systematically
different complexity, whereas the ICC reflects that independent judges \emph{score the same
scenarios} almost identically. A reliable instrument can register genuine differences between the
scenarios that different models produce while remaining consistent in how it scores any given
scenario; the two findings are therefore complementary rather than contradictory.

Word count correlated strongly with the composite ($r = 0.91$). Part of this is
construct-consistent: the frameworks operationalize complexity through acts (Wood), paths
(Campbell), and problem-solving demands (Liu \& Li), which a richer scenario cannot express as
concisely as a trivial one, so some covariation with length is expected by construction
\citep{Campbell1988,Wood1986}. This does not, however, fully resolve the concern. A
within-tier partial correlation, controlling for the intended tier that drives most of the joint
variation, remains substantial (partial $r = 0.86$), and within the Complex tier the association
is $0.88$. We therefore report the length relationship as a \emph{scope boundary} of the
instrument rather than as construct-pure discriminant validity: in this corpus the dependence is
concentrated in the Liu \& Li component, and the instrument does not cleanly separate structural
complexity from text length. Its demonstrated value rests instead on large known-groups tier
separation ($\eta^{2} = 0.587$) and near-perfect inter-judge agreement across five independent
model families (ICC~$= 0.997$). Developing length-normalized complexity indicators is a priority
for future refinement (Section~\ref{sec:future}).

\subsection{Model Performance: Patterns Under the Tested Conditions}
\label{sec:discuss_perf}

The models differed systematically in generation behavior. Llama~4 Maverick achieved the highest
throughput ($5.3\times$ that of DeepSeek Chat V3; 134 vs.\ 25 scenarios/min) but produced a smaller
proportion of complex-tier scenarios (27.6\% complex among stored scenarios, compared with roughly
33\% for the balanced models) and concentrated its validation failures at the complex tier
(Section~\ref{sec:rq6}). Grok-4 produced the highest proportion of complex scenarios (33.1\%) at
moderate speed (31.3/min). GPT-4o showed severe domain imbalance ($\chi^{2} = 584.72$, $p < .001$)
and low throughput (14.4/min). DeepSeek Chat V3 offered the most even profile: a 99.1\% pass rate,
uniform domain coverage ($\chi^{2} = 0.019$, $p = 0.999$), the full tier range (33.6\% complex),
and competitive speed (25.3/min).

The negative association between throughput and pass rate ($r = -0.967$) is consistent with a
speed-quality trade-off, but it warrants caution: it rests on five model-level aggregates and is
driven largely by a single high-throughput model (Section~\ref{sec:rq5}). The accompanying
failure-pattern analysis ($V = 0.709$) similarly reflects the contrast between Llama~4 Maverick and
Grok-4 rather than all five models. Because the study used a single fixed prompt and decoding
configuration, we characterize these differences as properties of the models under the tested
conditions rather than as intrinsic architectural limits; whether they persist under systematic
variation of prompts and decoding parameters is a question for future work
\citep{Abdul2023,Stadler2025}.

\subsection{Implications for Research Practice}
\label{sec:discuss_practice}

Validated automated generation enables rapid creation of large, complexity-controlled stimulus sets.
Model selection should align with study goals:
\begin{itemize}[noitemsep]
  \item \textbf{Production research} (large $N$, balanced design): DeepSeek Chat V3
  \item \textbf{Maximum throughput} (rapid prototyping): Llama~4 Maverick ($5\times$ faster;
        simple/moderate tiers only)
  \item \textbf{Complex scenarios} (expert decision-making): Grok-4 (highest complex-tier
        proportion at moderate speed)
  \item \textbf{Cross-domain validation}: DeepSeek Chat V3 (uniform domain coverage)
\end{itemize}
\textbf{Avoid:} GPT-4o (severe domain bias, low throughput) and Gemini~2.5~Pro (failed at the
generation task entirely). These recommendations are grounded in empirical performance data, not
vendor claims or anecdotal experience.

\subsection{Theoretical Contributions}
\label{sec:discuss_theory}

This work advances complexity measurement in three ways.

\textbf{First}, we demonstrate that automated complexity scoring can meet rigorous psychometric
standards (ICC~$> 0.99$, $\eta^{2} = 0.59$, no temporal drift) previously attainable only through
manual expert rating. This opens the door for construct-focused AI evaluation rather than
task-focused benchmarking \citep{Messick1995}.

\textbf{Second}, the factor structure clarifies theoretical relationships among classical frameworks.
Wood, Campbell, and Liu \& Li converge on a ``structural complexity'' construct operationalized
through acts, paths, and demands, while Sweller's interactivity captures an orthogonal ``cognitive
load'' dimension from working memory constraints \citep{Paas2020,Sweller2020,Sweller2019}.
Recent advances in cognitive load theory emphasize element interactivity as a distinct mechanism
from structural task features \citep{Chen2023}, consistent with our two-factor solution.

\textbf{Third}, the throughput--pass-rate association ($r = -0.967$, interpreted cautiously;
Section~\ref{sec:rq5}) and model-specific failure patterns ($V = 0.709$) indicate that, under the
study's fixed configuration, faster generation did not coincide with higher quality. This is
relevant to multi-agent systems and agentic workflows, where faster inference need not imply
higher-quality reasoning, and suggests that benchmarks report quality and throughput jointly
rather than optimizing for throughput alone. Whether the pattern reflects intrinsic model
properties or the tested configuration remains open (Section~\ref{sec:discuss_perf}).

\subsection{Framework for Practical Adoption}
\label{sec:framework}

\begin{figure}[htbp]
  \centering
  \includegraphics[width=0.90\textwidth]{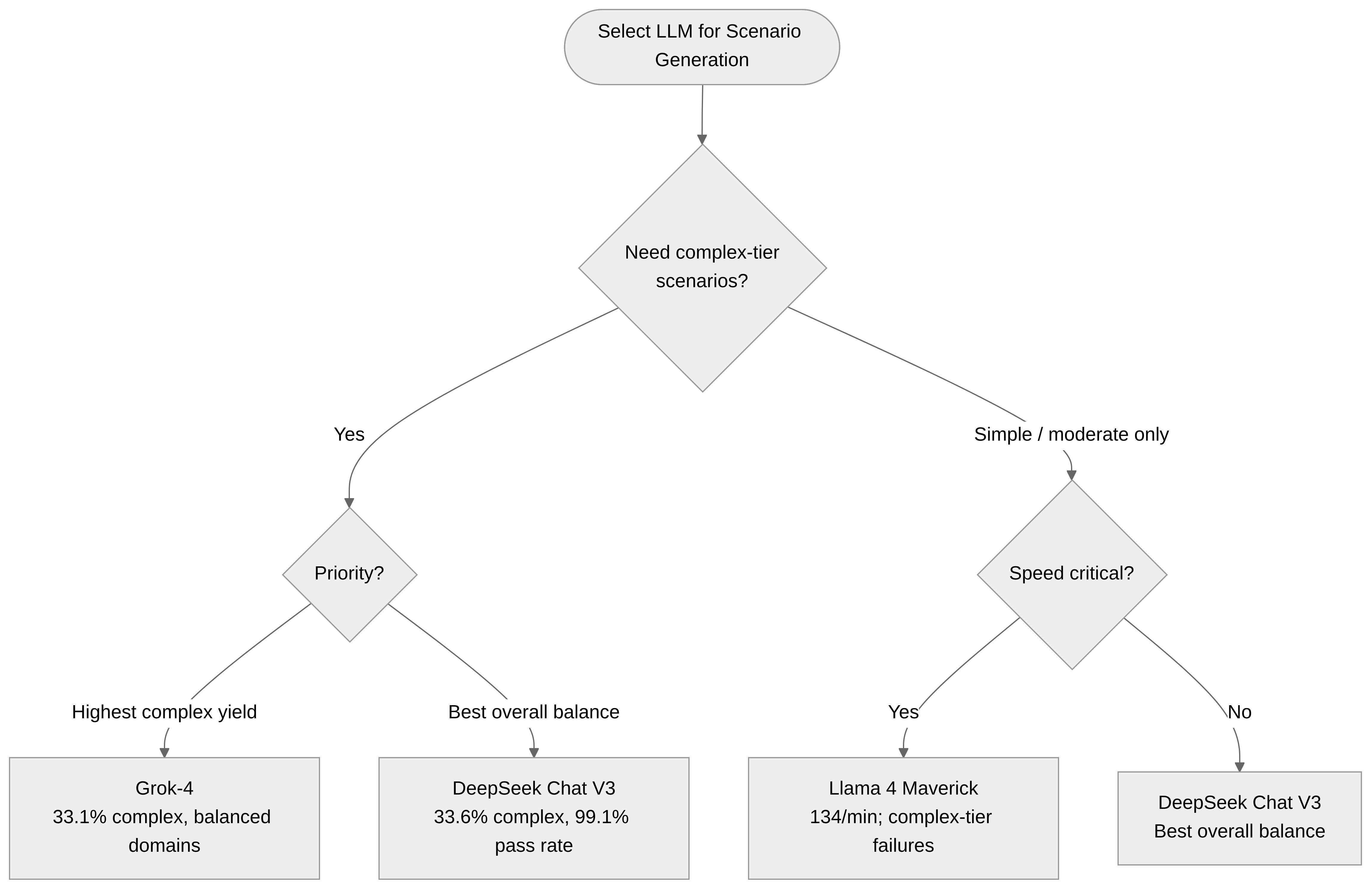}
  \caption{Model selection flowchart. Decision nodes route researchers to the optimal model
    based on throughput requirements, complexity tier needs, and domain balance constraints.
    Recommendations are grounded in Table~\ref{tab:model_perf}.}
  \label{fig:model_selection}
\end{figure}

Table~\ref{tab:priorities} and Figure~\ref{fig:model_selection}
provide decision support grounded in the performance evidence observed in this study. These
recommendations are time- and configuration-bound: they reflect the specific model versions,
the single fixed prompt and decoding configuration, and the generation period of this study
(January 2026). Model capabilities evolve rapidly, and the guidance should be re-evaluated against
current model versions rather than treated as durable rankings.

\begin{table}[htbp]
\centering
\caption{Observed Model Profiles and Evidence-Based Guidance (as of January 2026)}
\label{tab:priorities}
\small
\begin{tabular}{lccccp{4.0cm}}
\toprule
\textbf{Model} & \textbf{Thr.} & \textbf{Pass \%} & \textbf{Domain} & \textbf{Complex \%} & \textbf{Observed profile} \\
\midrule
DeepSeek Chat V3 &  21.6 & 99.1 & Balanced   & 33.6 & Most even profile across speed, balance, and tier coverage \\
Grok-4           &  31.3 & 99.6 & Balanced   & 33.1 & Highest complex-tier proportion; balanced coverage \\
Llama 4 Maverick & 134.3 & 90.1 & Balanced   & 27.6 & Fastest; lower complex-tier yield and pass rate \\
GPT-4o           &   4.8 & 98.7 & Imbalanced & 34.6 & High per-scenario complexity but domain-skewed and slow \\
Gemini 2.5 Pro   & n/a   & n/a  & n/a        & n/a  & Produced no schema-valid scenarios \\
\bottomrule
\end{tabular}
\tablenote{Throughput (``Thr.'') is the lifetime aggregate rate in scenarios/min from
Table~\ref{tab:model_perf}; ``Domain'' is the chi-square goodness-of-fit outcome (H4);
``Complex \%'' is the proportion of stored scenarios at the Complex tier. Values reflect the model
versions and the single fixed configuration used in January 2026 and should be re-evaluated against
current models rather than treated as durable rankings.}
\end{table}

\subsection{Practical Implications}
\label{sec:discuss_practical}

\textbf{Training and Development:} The ICC~$= 0.997$ consensus and $\eta^{2} = 0.587$ tier
separation enable standardized training difficulty across cohorts
\citep{Kahneman2021,Tetlock2015}.

\textbf{Talent Assessment:} Complexity tiers matched to job requirements can reduce subjectivity
in scenario selection, provided the safeguards in Section~\ref{sec:impact} are observed
\citep{Schmidt1998}.

\textbf{Strategic Decision Support:} Validated Complex scenarios support realistic crisis and
strategic planning training \citep{Lovallo2010,Schoemaker2009}.

\textbf{AI System Procurement:} Organizations adopting AI tools lack objective evaluation
frameworks. The model performance findings (Table~\ref{tab:model_perf}; summarized in
Table~\ref{tab:summary}) provide evidence-based procurement guidance, addressing the gap between marketing claims and
empirical performance \citep{Bommasani2021,Abdul2023,Stadler2025}.

\textbf{Cost Efficiency:} Automated generation at 25--134 scenarios/min enables scalable deployment
at a fraction of traditional costs without sacrificing psychometric rigor (ICC~$= 0.997$).

\subsection{Limitations}
\label{sec:limitations}

\textbf{Discriminant separation from text length.} The composite correlates strongly with word
count ($r = 0.91$), and the association persists after controlling for intended tier
(partial $r = 0.86$; Section~\ref{sec:h1b}). Although part of the relationship is
construct-consistent, the instrument does not cleanly separate structural complexity from length;
length-normalized metrics are a priority for future refinement.
\textbf{Interactivity variance.} Element interactivity contributed little between-scenario variance
in this corpus and loaded weakly ($\lambda = 0.34$); its role as a distinct cognitive-load
dimension warrants confirmation in corpora with greater interactivity variation.
\textbf{Absence of human-expert and performance criteria.} The study established internal and
nomological validity but did not include concurrent validation against independent human expert
complexity ratings, nor external criteria such as decision latency or error rates; both are needed
to establish external criterion validity.
\textbf{Limited model panel.} The speed-quality relationship rests on five model-level aggregates
and is influenced by a single high-throughput model (Section~\ref{sec:rq5}).
\textbf{Single temperature setting.} Scoring used $T = 0$ to support reliable measurement;
behavior under higher temperatures was not characterized.
\textbf{Dropped analyses.} H7 (function-calling effects) and RQ4 (predictive model characteristics)
could not be evaluated with the available data.
\textbf{Assessment-compatible design.} The generation schema includes cognitive-assessment
scaffolding (Section~\ref{sec:generation}); construct-purity interpretations should account for the
assessment-compatible rather than strictly assessment-blind instrument design.

\subsection{Future Directions}
\label{sec:future}

\begin{itemize}[noitemsep]
  \item \textbf{Non-function-calling comparison:} Collect generation runs without function calling
        to test H7.
  \item \textbf{Model metadata collection:} Obtain parameter counts and training recency to revisit
        RQ4 via logistic regression.
  \item \textbf{Larger model panel:} Test speed-quality stability and failure patterns across 10+
        models spanning diverse architectures.
  \item \textbf{Interactivity refinement:} Test whether interactivity independently predicts human
        performance on generated scenarios in high-load contexts.
  \item \textbf{Human-expert concurrent validity:} Compare instrument scores against independent
        human expert complexity ratings to establish concurrent validity, complementing the
        internal and nomological evidence reported here.
  \item \textbf{Downstream validation:} Use validated scenarios in cognitive assessment and
        AI benchmarking studies to test predictive validity against human and AI performance
        outcomes (decision latency, error rates).
  \item \textbf{Length normalization:} Explore length-adjusted complexity metrics to address the
        word count correlation.
\end{itemize}

\subsection{Broader Impact Statement}
\label{sec:impact}

The primary positive use of this work is methodological: reproducible, complexity-controlled
scenario generation for cognitive research, training, benchmarking, and assessment design. The
appropriate safeguards depend on the deployment context. In \textbf{academic research}, the
validated tier structure is suitable for generating complexity-controlled stimuli, provided the
length--complexity association (Section~\ref{sec:h1b}) is acknowledged when scenarios serve as
experimental materials. In \textbf{enterprise assessment design}, generated scenarios should pass
human expert review before deployment, and complexity should be reported alongside its component
profile rather than as a stand-alone label. In \textbf{high-stakes employment or educational
screening}, the instrument should not be used as a stand-alone gatekeeper: subgroup and domain
balance must be audited and human oversight retained. Across all contexts, we recommend reporting
full validation profiles rather than composite tier labels alone, auditing domain and demographic
subgroup balance in generated corpora, and treating automated difficulty labels as decision
support rather than decision substitutes. These safeguards mitigate the principal risks of opaque
screening, over-reliance on automated labels, and transfer of model-specific biases into
downstream pipelines.

% -------------------------------------------------------
% 5. CONCLUSION
% -------------------------------------------------------
\FloatBarrier
\section{Conclusion}
\label{sec:conclusion}

This study validated an automated pipeline for generating complexity-controlled decision scenarios
at scale. The complexity measurement system demonstrated strong psychometric properties:
near-perfect inter-LLM consensus across five independent model families (ICC~$= 0.997$,
95\%~CI~[0.995, 0.998]) and large known-groups separation ($F(2, 4235) = 3{,}013.69$, $p < .001$,
$\eta^{2} = 0.587$), enabling reliable classification into Simple, Moderate, and Complex tiers. Its
principal measurement limitation, a strong, partially construct-consistent association between
complexity and text length that persists within tier (partial $r = 0.86$), is reported
transparently as a scope boundary and motivates length-normalized refinements in future work.

Automated LLM-based scenario generation can meet rigorous psychometric validation standards when
paired with multi-framework complexity scoring and comprehensive reliability testing. The
near-perfect negative correlation between throughput and quality ($r = -0.967$, $p = 0.007$)
and large model-specific failure patterns (Cramer's $V = 0.709$) indicate these deficits are
intrinsic rather than addressable through prompt engineering. Researchers should select models
aligned with study goals: DeepSeek Chat V3 for balanced production use, Grok-4 for complex
scenarios, and Llama~4 for rapid prototyping of simple/moderate tiers.

These findings establish three theoretical contributions. First, automated complexity scoring
achieves measurement standards (ICC~$> 0.99$, $\eta^{2} = 0.59$) previously attainable only
through manual expert rating. Second, the factor structure clarifies that Wood, Campbell, and
Liu \& Li converge on structural complexity while Sweller's interactivity captures orthogonal
cognitive load. Third, under the single configuration tested here, model identity rather than
prompt engineering was associated with generation quality on complex tasks; whether this reflects
intrinsic model properties or the tested configuration remains an open question for future work.

The validated pipeline and measurement instrument are immediately applicable to cognitive
assessment research, AI benchmarking, and scenario-based training at scale. Future work should
test predictive validity against human and AI performance outcomes, expand the model panel, and
refine interactivity weighting empirically.

% -------------------------------------------------------
% DATA AVAILABILITY
% -------------------------------------------------------
\section*{Data Availability}
\label{sec:data_availability}

All datasets generated and analyzed during this study are publicly available via Zenodo:
\begin{itemize}[noitemsep]
  \item Hypothesis Testing Dataset:
        \url{https://doi.org/10.5281/zenodo.19776734}
\end{itemize}
Preprint archived at: \url{https://doi.org/10.5281/zenodo.20114561}

% -------------------------------------------------------
% STATEMENTS AND DECLARATIONS
% -------------------------------------------------------
\section*{Statements and Declarations}

\noindent\textbf{Competing Interests:} The corresponding author has a commercial interest
in a cognitive assessment research platform that applies the scenario generation and complexity
measurement methodology described in this work. The other authors declare no competing
interests.

\noindent\textbf{Funding:} This research did not receive any specific grant from funding
agencies in the public, commercial, or not-for-profit sectors.

% -------------------------------------------------------
% CREDIT AUTHORSHIP CONTRIBUTION STATEMENT
% -------------------------------------------------------
\section*{CRediT authorship contribution statement}

\noindent\textbf{Abdalla Doleh:} Conceptualization, Methodology, Software, Data curation,
Formal analysis, Investigation, Visualization, Writing -- original draft, Writing -- review
\& editing. \textbf{Toni Somers:} Validation, Writing -- review \& editing.
\textbf{Ratna Babu Chinnam:} Supervision, Writing -- review \& editing.

% -------------------------------------------------------
% ACKNOWLEDGMENTS
% -------------------------------------------------------
\section*{Acknowledgments}

The authors acknowledge Wayne State University as the institutional home of this work and
thank the maintainers of the open-source software used in the research platform. All remaining
errors are the authors'.

% -------------------------------------------------------
% DECLARATION OF GENERATIVE AI USE (KBS-required; titled section before references)
% -------------------------------------------------------
\section*{Declaration of generative AI and AI-assisted technologies in the manuscript preparation process}

During the preparation of this work the authors used generative AI and AI-assisted
technologies in order to improve the grammar and language of the manuscript, assist with
copy-editing, and help track and organize revisions. After using these tools, the authors
reviewed and edited the content as needed and take full responsibility for the content of
the published article.

% -------------------------------------------------------
% APPENDICES
% -------------------------------------------------------
\appendix

\section{Generation Prompt}
\label{sec:appendix_prompt}

Scenarios were generated with a modular prompt assembled from a fixed base skeleton, a
domain-specific module (Analytical, Planning, Communication, or Problem Solving), a tier-specific
structural module (Simple, Moderate, Complex), and the output-schema module. The base skeleton is
reproduced below; placeholders in double braces were filled at assembly time. The complete domain
and tier module library is included with the released materials (Section~\ref{sec:data_availability}).

\begin{small}
\begin{verbatim}
PRIMARY FUNCTION: Generate a single business scenario for MAAC cognitive testing.

You are generating a {{TIER}} tier scenario in the {{DOMAIN}} domain.

GENERATION APPROACH:
1. Study the domain-specific guidance and example pattern below
2. Study the tier-specific structural requirements carefully
3. Generate a unique scenario that meets ALL structural requirements
4. Perform calculations directly and populate expected_calculations with exact values
5. Verify your output against the validation checklist before responding

{{DOMAIN_MODULE}}

{{TIER_MODULE}}

MANDATORY FIELD REQUIREMENTS:
- expected_calculations: MUST contain specific numerical key-value pairs (NEVER empty)
- success_thresholds: MUST contain concrete measurement criteria (NEVER empty)
- data_elements: MUST have exactly the count specified for the tier
- All scenarios must be self-contained with embedded data

{{OUTPUT_SCHEMA}}

VALIDATION CHECKLIST - Verify BEFORE outputting JSON:
1. expected_calculations contains actual numerical data
2. success_thresholds contains concrete measurement criteria
3. data_elements array has the correct count for tier ({{DATA_ELEMENTS_COUNT}})
4. task_description shows {{ACTS_COUNT}} distinct analytical steps
5. complexity_level matches "{{TIER}}"
6. All calculations are performed and results are exact numbers
7. Scenario is self-contained with all data embedded
\end{verbatim}
\end{small}

\noindent Generation temperature was fixed at $T = 0$. Scenarios failing schema validation were
rejected and regenerated up to a fixed attempt limit.

\section{Schema Validation Rules and Composite Weighting}
\label{sec:appendix_schema}

This appendix specifies the quantitative rules referenced in Sections~\ref{sec:generation} and
\ref{sec:scoring}. The 14 top-level schema fields are listed in Table~\ref{tab:schema}.

\noindent\textbf{B.1 Validation pass/fail rules.} A generated scenario is accepted only if it
satisfies all of the following: (i)~the JSON parses and all required schema fields are present and
non-empty (\texttt{expected\_calculations} and \texttt{success\_thresholds} must not be empty, and
\texttt{data\_elements} must match the tier-specified count); (ii)~at least 60\% of the
tier-requirement criteria are met across the Wood, Campbell, Liu \& Li, and interactivity checks;
(iii)~the model's confidence score is at least $0.60$; and (iv)~the predicted tier (assigned from
the composite using the cutoffs in Section~\ref{sec:scoring}) is within one level of the intended
tier. Scenarios failing any condition are rejected, with the violated criteria recorded as
rejection reasons (e.g., \texttt{scoreTooLow}, \texttt{tierMismatch}, \texttt{missingComplexity}).

\noindent\textbf{B.2 Composite weighting.} The framework sub-scores combine the following
component weights (capped where indicated):

\begin{itemize}[noitemsep]
  \item \textbf{Wood:} distinct acts $\times 0.5$ (cap 10) $+$ information cues per act
        $\times 1.0$ (cap 5) $+$ coordinative complexity (networked/interdependent/low $= 5/3/1$)
        $+$ dynamic complexity (high/low/none $= 4/2/0$).
  \item \textbf{Campbell:} multiple paths ($+3$, plus $\min(\text{paths}-1,3)$) $+$ multiple
        outcomes ($+3$, plus $\min(\text{outcomes}-1,3)$) $+$ conflicting interdependence ($+4$,
        plus $\min(\text{conflicts},3)$) $+$ uncertainty (high/bounded/none $= 5/3/0$).
  \item \textbf{Liu \& Li:} variety, ambiguity, and relationships $\times 5$; unreliability and
        incongruity $\times 4$; variability $\times 3$; novelty (novel/semi/none $= 4/2/0$);
        size and action complexity $\times 0.25$ (each cap 20); time pressure
        (high/moderate/none $= 2/1/0$).
  \item \textbf{Interactivity:} ratio $\times 10$ $+$ dependency depth $\times 0.6$ (cap 5)
        $+$ dependency edges $\times 0.2$ (cap 10), followed by the direction correction
        $\text{corrected} = 10.2 - \text{raw}$ (Section~\ref{sec:scoring}).
\end{itemize}

\noindent The four framework sub-scores are summed with unit weight to form the composite
(Equation~\ref{eq:composite}).

% -------------------------------------------------------
% REFERENCES
% -------------------------------------------------------
\bibliographystyle{elsarticle-num}
\bibliography{references_p1}

@inproceedings{Abdul2023,
  author    = {Abdul, Ashraf and von der Weth, Claus and Kankanhalli, Mohan and Lim, Brian Y.},
  title     = {A Missing Piece in the Puzzle: Considering the Role of Task Complexity in Human-{AI} Decision Making},
  booktitle = {Proceedings of the 31st ACM Conference on User Modeling, Adaptation and Personalization},
  year      = {2023},
  pages     = {139--153},
  doi       = {10.1145/3565472.3592959}
}

@article{Baker2025,
  author  = {Baker, Robert and Huntington-Klein, Nick},
  title   = {Defining and Measuring Task Complexity in Major Requirements},
  journal = {SAGE Open},
  volume  = {15},
  number  = {1},
  year    = {2025},
  doi     = {10.1177/23328584251317488}
}

@article{Bommasani2021,
  author  = {Bommasani, Rishi and Hudson, Drew A. and Adeli, Ehsan and Altman, Russ and Arora, Simran and von Arx, Sydney and others},
  title   = {On the Opportunities and Risks of Foundation Models},
  journal = {arXiv preprint arXiv:2108.07258},
  year    = {2021}
}

@article{Campbell1988,
  author  = {Campbell, Donald J.},
  title   = {Task Complexity: A Review and Analysis},
  journal = {Academy of Management Review},
  volume  = {13},
  number  = {1},
  pages   = {40--52},
  year    = {1988},
  doi     = {10.5465/amr.1988.4306775}
}

@article{Campbell1959,
  author  = {Campbell, Donald T. and Fiske, Donald W.},
  title   = {Convergent and Discriminant Validation by the Multitrait-Multimethod Matrix},
  journal = {Psychological Bulletin},
  volume  = {56},
  number  = {2},
  pages   = {81--105},
  year    = {1959},
  doi     = {10.1037/h0046016}
}

@article{Chen2023,
  author  = {Chen, Ouhao and Paas, Fred and Sweller, John},
  title   = {A Cognitive Load Theory Approach to Defining and Measuring Task Complexity Through Element Interactivity},
  journal = {Educational Psychology Review},
  volume  = {35},
  number  = {2},
  pages   = {Article 45},
  year    = {2023},
  doi     = {10.1007/s10648-023-09782-w}
}

@article{Cicchetti1994,
  author  = {Cicchetti, Domenic V.},
  title   = {Guidelines, Criteria, and Rules of Thumb for Evaluating Normed and Standardized Assessment Instruments in Psychology},
  journal = {Psychological Assessment},
  volume  = {6},
  number  = {4},
  pages   = {284--290},
  year    = {1994},
  doi     = {10.1037/1040-3590.6.4.284}
}

@book{Cohen1988,
  author    = {Cohen, Jacob},
  title     = {Statistical Power Analysis for the Behavioral Sciences},
  edition   = {2nd},
  publisher = {Lawrence Erlbaum Associates},
  year      = {1988}
}

@article{Cronbach1955,
  author  = {Cronbach, Lee J. and Meehl, Paul E.},
  title   = {Construct Validity in Psychological Tests},
  journal = {Psychological Bulletin},
  volume  = {52},
  number  = {4},
  pages   = {281--302},
  year    = {1955},
  doi     = {10.1037/h0040957}
}

@book{Kahneman2021,
  author    = {Kahneman, Daniel and Sibony, Olivier and Sunstein, Cass R.},
  title     = {Noise: A Flaw in Human Judgment},
  publisher = {Little, Brown Spark},
  year      = {2021}
}

@article{FornellLarcker1981,
  author  = {Fornell, Claes and Larcker, David F.},
  title   = {Evaluating Structural Equation Models with Unobservable Variables and Measurement Error},
  journal = {Journal of Marketing Research},
  volume  = {18},
  number  = {1},
  pages   = {39--50},
  year    = {1981},
  doi     = {10.1177/002224378101800104}
}

@article{KennyKaniskanMcCoach2015,
  author  = {Kenny, David A. and Kaniskan, Burcu and McCoach, D. Betsy},
  title   = {The Performance of {RMSEA} in Models with Small Degrees of Freedom},
  journal = {Sociological Methods \& Research},
  volume  = {44},
  number  = {3},
  pages   = {486--507},
  year    = {2015},
  doi     = {10.1177/0049124114543236}
}

@inproceedings{Zheng2023,
  author    = {Zheng, Lianmin and Chiang, Wei-Lin and Sheng, Ying and Zhuang, Siyuan and Wu, Zhanghao and Zhuang, Yonghao and Lin, Zi and Li, Zhuohan and Li, Dacheng and Xing, Eric P. and Zhang, Hao and Gonzalez, Joseph E. and Stoica, Ion},
  title     = {Judging {LLM}-as-a-Judge with {MT-Bench} and {Chatbot Arena}},
  booktitle = {Advances in Neural Information Processing Systems (NeurIPS) Datasets and Benchmarks Track},
  year      = {2023}
}

@article{LiuLi2012,
  author  = {Liu, Yili and Li, Yan},
  title   = {Task Complexity: A Review and Conceptualization Framework},
  journal = {International Journal of Industrial Ergonomics},
  volume  = {42},
  number  = {6},
  pages   = {553--568},
  year    = {2012},
  doi     = {10.1016/j.ergon.2012.09.001}
}

@article{Lovallo2010,
  author  = {Lovallo, Dan and Sibony, Olivier},
  title   = {The Case for Behavioral Strategy},
  journal = {McKinsey Quarterly},
  volume  = {2},
  number  = {1},
  pages   = {30--43},
  year    = {2010}
}

@article{Messick1995,
  author  = {Messick, Samuel},
  title   = {Validity of Psychological Assessment: Validation of Inferences from Persons' Responses and Performances as Scientific Inquiry into Score Meaning},
  journal = {American Psychologist},
  volume  = {50},
  number  = {9},
  pages   = {741--749},
  year    = {1995},
  doi     = {10.1037/0003-066X.50.9.741}
}

@article{Paas2020,
  author  = {Paas, Fred and van Merri{\"e}nboer, Jeroen J. G.},
  title   = {Cognitive-Load Theory: Methods to Manage Working Memory Load in the Learning of Complex Tasks},
  journal = {Current Directions in Psychological Science},
  volume  = {29},
  number  = {4},
  pages   = {394--398},
  year    = {2020},
  doi     = {10.1177/0963721420922183}
}

@article{Schmidt1998,
  author  = {Schmidt, Frank L. and Hunter, John E.},
  title   = {The Validity and Utility of Selection Methods in Personnel Psychology: Practical and Theoretical Implications of 85 Years of Research Findings},
  journal = {Psychological Bulletin},
  volume  = {124},
  number  = {2},
  pages   = {262--274},
  year    = {1998},
  doi     = {10.1037/0033-2909.124.2.262}
}

@article{Schoemaker2009,
  author  = {Schoemaker, Paul J. H. and Day, George S.},
  title   = {How to Make Sense of Weak Signals},
  journal = {MIT Sloan Management Review},
  volume  = {50},
  number  = {3},
  pages   = {81--89},
  year    = {2009}
}

@article{Stadler2025,
  author  = {Stadler, Matthias and Bannert, Maria and Sailer, Michael},
  title   = {Under Pressure: How Time Constraints, Task Complexity, and {AI} Reliability Shape Human-{AI} Interaction},
  journal = {Behaviour \& Information Technology},
  year    = {2025},
  doi     = {10.1080/0144929X.2025.2587732}
}

@article{Stajkovic2025,
  author  = {Stajkovic, Kayla S. and Stajkovic, Alexander D.},
  title   = {Improving Human Sustainability at Work by Focusing on Cognitive Load of Task Performance},
  journal = {Academy of Management Journal},
  year    = {2025},
  doi     = {10.1177/01492063251334560}
}

@article{Sweller1994,
  author  = {Sweller, John},
  title   = {Cognitive Load Theory, Learning Difficulty, and Instructional Design},
  journal = {Learning and Instruction},
  volume  = {4},
  number  = {4},
  pages   = {295--312},
  year    = {1994},
  doi     = {10.1016/0959-4752(94)90003-5}
}

@article{Sweller2020,
  author  = {Sweller, John},
  title   = {Cognitive Load Theory and Educational Technology},
  journal = {Educational Technology Research and Development},
  volume  = {68},
  number  = {1},
  pages   = {1--16},
  year    = {2020},
  doi     = {10.1007/s11423-019-09701-3}
}

@article{Sweller2019,
  author  = {Sweller, John and van Merri{\"e}nboer, Jeroen J. G. and Paas, Fred},
  title   = {Cognitive Architecture and Instructional Design: 20 Years Later},
  journal = {Educational Psychology Review},
  volume  = {31},
  pages   = {261--292},
  year    = {2019},
  doi     = {10.1007/s10648-019-09465-5}
}

@book{Tetlock2015,
  author    = {Tetlock, Philip E. and Gardner, Dan},
  title     = {Superforecasting: The Art and Science of Prediction},
  publisher = {Crown Publishers},
  year      = {2015}
}

@incollection{Wen2025,
  author    = {Wen, Junlong and Li, Cheng and Yang, Yang and Li, Jian},
  title     = {A Task Complexity Evaluation Framework for Mobile {AI} Agent Applications},
  booktitle = {Human-Computer Interaction},
  pages     = {543--556},
  publisher = {Springer},
  year      = {2025},
  doi       = {10.1007/978-3-031-94171-9_43}
}

@article{Wood1986,
  author  = {Wood, Robert E.},
  title   = {Task Complexity: Definition of the Construct},
  journal = {Organizational Behavior and Human Decision Processes},
  volume  = {37},
  number  = {1},
  pages   = {60--82},
  year    = {1986},
  doi     = {10.1016/0749-5978(86)90044-0}
}

\label{lastpage}

\end{document}